\documentclass[letterpaper, 10 pt, conference]{ieeeconf}
\IEEEoverridecommandlockouts
\makeatletter
\let\NAT@parse\undefined
\makeatother
\usepackage[colorinlistoftodos]{todonotes}
\usepackage{cite}
\usepackage{graphicx}
\usepackage{times}
\usepackage{amsmath}
\usepackage{amssymb}
\usepackage{booktabs}
\usepackage{array}
\usepackage{xcolor}
\usepackage[utf8]{inputenc}
\usepackage[pdfstartview=FitH,bookmarksnumbered=true,colorlinks,linkcolor=black,citecolor=black,bookmarksopen=true]{hyperref}

\title{\LARGE \bf
Gated Residual Body--Hand Coordination\\
for Whole-Body Humanoid Teleoperation
}

\newif\ificraanonymous
\icraanonymousfalse

\ificraanonymous
  \author{Anonymous Authors}
\else
  \author{
    Ruiming Wu$^{1,2}$,
    Shuang Li$^{2}$,
    Liding Zhang$^{1,\dagger}$,
    Alois Knoll$^{1}$,
    Zhaopeng Chen$^{2}$%
    \thanks{%
      $^{1}$ Informatics 6 -- Chair of Robotics, Artificial Intelligence and Real-time Systems,
      Technical University of Munich, Munich, Germany. }
  \thanks{$^{2}$ Agile Robots SE, Munich, Germany. } 
      \thanks{$^{\dagger}$\text{Corresponding author: Liding Zhang}(\texttt{liding.zhang@tum.de}).
    }%
  }
\fi

\IEEEaftertitletext{%
  \vspace{-0.6\baselineskip}%
  \noindent
  \includegraphics[width=\textwidth]{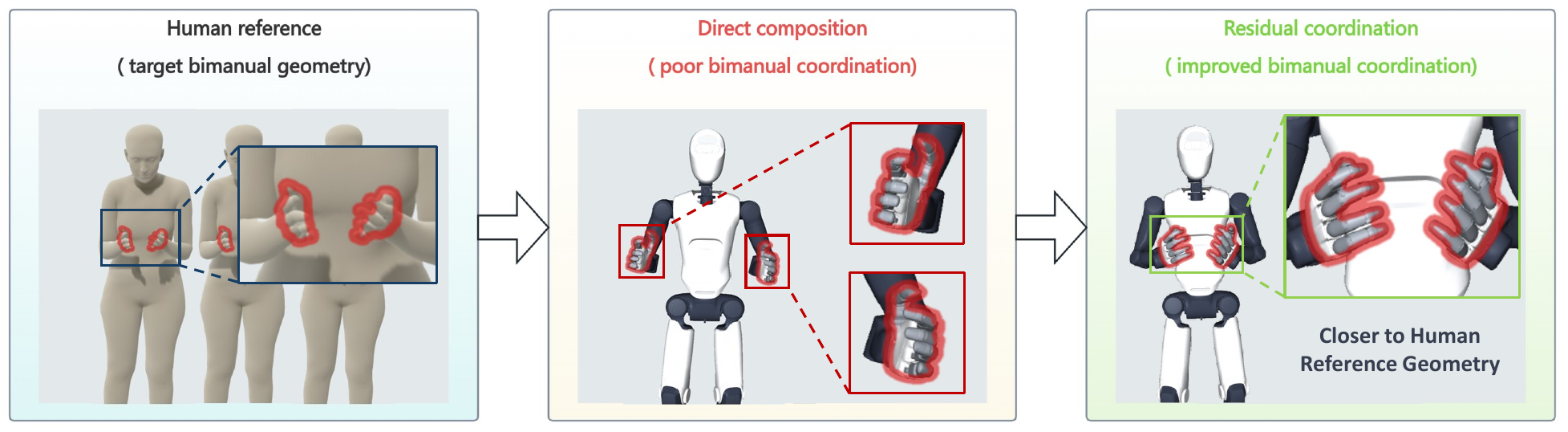}\\[0.15em]%
  \refstepcounter{figure}\label{fig:first-page-qualitative}%
  \noindent
  \parbox{\textwidth}{%
  \footnotesize\raggedright
  Fig.~\thefigure. Qualitative comparison of body--hand coordination for the
  same human reference. Direct composition of the frozen whole-body tracker
  and dexterous-hand command generator produces a mismatch in relative wrist
  and hand geometry (center), whereas the proposed residual coordination more
  closely reproduces the target bimanual configuration (right). Red contours
  highlight the hands.
}%
  \vspace{0.2\baselineskip}%
}

\begin{document}
\bstctlcite{IEEEtran:BSTcontrol}

\maketitle
\thispagestyle{empty}
\pagestyle{empty}

\begin{abstract}
Whole-body humanoid teleoperation commonly combines a whole-body motion-tracking policy with a separate retargeting module for multifingered hands. 
However, direct composition does not explicitly preserve geometric relations across the body and hand streams, allowing errors in wrist placement and finger retargeting to distort bimanual interaction geometry.
To establish a reliable nominal body stream on the target humanoid, we develop a multi-pose morphology-calibration method that jointly estimates triaxial scales and effector offsets, together with staged motion dataset curation for whole-body tracker training. 
We then introduce a gated residual coordination framework that keeps the body tracker and hand retargeter frozen while applying bounded corrections to their commands. Reference-geometry-dependent reward gates weight interaction objectives according to their relevance, while a motion-conditioned action gate adjusts correction authority according to the current motion context.

On held-out interaction motions, our method reduces wrist and fingertip geometry errors by 39.2--56.3\% over direct composition, while maintaining comparable whole-body tracking ability on diverse motions (89.03\% without coordination and 89.29\% with it). Ablations characterize the contributions of gated objectives, adaptive correction authority, and joint-group-specific residual control.

\end{abstract}

\section{Introduction}

Recent humanoid tracking policies can reproduce a broad range of human motion
from proprioceptive observations and motion-reference streams
\cite{he2024h2o,he2024omnih2o,luo2025sonic}. However, these policies do not directly control the five-fingered hand; instead, finger commands are usually generated by a separate hand-retargeting module\cite{handa2019dexpilot}. 
Such a modular decomposition improves system flexibility and reusability by allowing the body and hand components to be designed, validated, and updated independently.
Even when each module tracks its own reference well, their direct composition does not explicitly preserve geometric relations across the two kinematic chains. Small errors in wrist placement and wrist-local finger retargeting can therefore accumulate, producing inaccurate relative wrist poses and fingertip positions during bimanual interaction.

Figure~\ref{fig:first-page-qualitative} illustrates this mismatch
and the improved bimanual configuration achieved by the proposed
residual coordination.

A natural response is to optimize the body and hand commands jointly so that
cross-stream geometric relations can be enforced explicitly. 
Realizing this
idea on a new humanoid embodiment, however, presents two challenges. 
First, the nominal body stream is not directly transferable across embodiments: tracker training requires robot-specific motion references, yet single-pose morphology calibration cannot uniquely distinguish body scaling from effector-local offsets.
Moreover, large retargeted motion collections may contain geometrically invalid poses, physical and temporal artifacts, and substantial redundancy \cite{guan2026limmt}. 
Second, replacing the modular body--hand stack with a monolithic controller would enlarge the action and optimization spaces, couple the development of the two subsystems, and require joint retraining whenever either component is updated.

We address the first challenge with multi-pose morphology calibration and staged motion dataset curation. The calibration jointly estimates root-frame triaxial scales and effector-local offsets from multiple paired human--robot poses, resolving the ambiguity of single-pose fitting. The retargeted motions are subsequently subjected to geometric validation, physical-quality screening, temporal discontinuity repair, ground-contact alignment, and density-aware selection. Applied to the large-scale SEED dataset \cite{bonesstudio2026seed}, this process produces robot-specific references for training a SONIC-based nominal whole-body tracker \cite{luo2025sonic}. A DexPilot-style geometric retargeter \cite{handa2019dexpilot} provides the nominal finger commands. The resulting body and hand command streams define the direct-composition baseline that our residual policy subsequently coordinates.

To address the second challenge, we introduce a residual policy
that jointly refines the frozen body and hand commands through separate,
joint-group-bounded corrections. Residual reinforcement learning retains the
capabilities of existing controllers while restricting learning to corrective
actions \cite{silver2018residual,johannink2019residual}. In our setting, residual coordination must determine not only how much correction to
apply, but also where and when to apply it. 
We consequently introduce two complementary gating mechanisms. A
motion-conditioned action gate allocates residual correction authority across
locomotion and manipulation joint groups according to the current human-motion
context. During training, reference-geometry-dependent reward gates emphasize
interaction objectives only when the corresponding geometric relations are
relevant. 

The paper makes three contributions:

\begin{itemize}

\item a residual coordination framework that preserves a
modular whole-body tracker and hand retargeter while jointly refining their
commands through separate, joint-group-bounded body and hand corrections;

\item two complementary context mechanisms: reference-geometry-dependent
reward gating that localizes interaction supervision during training, and
motion-conditioned action gating that allocates residual correction authority
during execution;

\item a multi-pose morphology-calibration method that resolves the scale--offset ambiguity of single-pose fitting by jointly estimating root-frame triaxial scales and effector-local offsets, together with staged motion dataset curation for whole-body tracker training.
\end{itemize}

\section{Related Work}

\subsection{Motion Data, Retargeting, and Curation}

Large-scale generalist humanoid tracking requires human motion to be converted
into embodiment-specific robot references, typically through optimization- or
learning-based retargeting~\cite{zhang2025mit}. General Motion Retargeting (GMR) provides a general
cross-embodiment formulation and emphasizes the effect of retargeting quality
on downstream tracking \cite{araujo2025gmr}. NVIDIA SOMA Retargeter provides a
practical pipeline for mapping SOMA-format motion to configured humanoid
embodiments \cite{nvidia2026somaretargeter}. Interaction-oriented methods such
as OmniRetarget extend pose correspondence by preserving
human--object--environment relations when such information is available
\cite{yang2025omniretarget}. Neural Motion Retargeting (NMR) further addresses
temporal failures in optimization-based retargeting, showing that acceptable
frame-wise solutions may still contain abrupt joint transitions and other
trajectory-level artifacts \cite{zhao2026nmr}.

At corpus scale, reference preparation also becomes a data-curation problem.
Human-to-Humanoid (H2O) uses a privileged policy to identify robot-feasible
references \cite{he2024h2o}, while Advanced Expressive Whole-Body Control
(ExBody2) filters motions according to downstream tracking feasibility
\cite{ji2024exbody2}. Less Is More for Motion Tracking (LIMMT) treats curation
itself as a central component of motion-tracker training, characterizing
motion data through physical feasibility, diversity, and complexity and
showing that a curated subset can outperform a larger unfiltered corpus
\cite{guan2026limmt}.

\subsection{Whole-Body Tracking and Dexterous Teleoperation}

Recent humanoid systems have expanded the coverage and deployability of
whole-body tracking. H2O and Omni Human-to-Humanoid (OmniH2O) learn deployable
control from human references and proprioception
\cite{he2024h2o,he2024omnih2o}, while General Motion Tracking (GMT) and
BeyondMimic target broader generalist tracking and downstream control
\cite{chen2025gmt,liao2025beyondmimic}. Supersizing Motion Tracking for Natural
Humanoid Control (SONIC) further scales tracking through a shared
representation across human and robot motion inputs \cite{luo2025sonic}.

Dexterous teleoperation introduces a separate human-to-robot correspondence
problem at the hand. DexPilot preserves grasp-relevant geometric relations
rather than directly matching anatomical joint angles
\cite{handa2019dexpilot}, while AnyTeleop generalizes optimization-based
retargeting across different arm--hand configurations
\cite{qin2023anyteleop}. HumDex instead learns a lightweight runtime mapping
from human fingertip observations to robot-hand targets
\cite{heng2026humdex}. Whole-body systems such as OmniH2O and HumDex combine
such hand commands with learned body controllers through modular control
paths \cite{he2024omnih2o,heng2026humdex}. WristMimic instead couples body and
wrist references with object-tracking objectives and contact dynamics
\cite{yu2026wristmimic}. These systems therefore span modular body--hand
composition and interaction-aware coupling, while command-level coordination
between independently generated body and hand commands remains a distinct
setting.

\subsection{Residual Coordination and Gating}
\begin{figure*}[t]
  \centering
  \includegraphics[width=\textwidth]
    {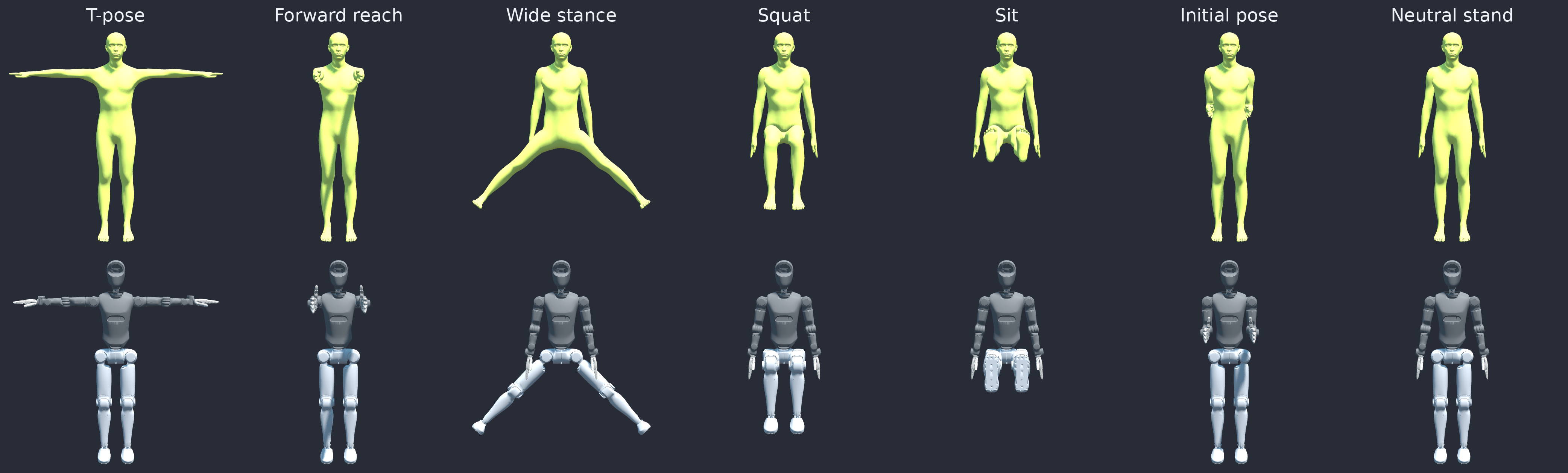}
  \caption{Seven manually paired poses for morphology calibration.
  Human reference poses (top) are paired with Agile One poses (bottom)
  to jointly estimate root-frame triaxial scales and effector-local offsets.}
  \label{fig:multi-pose-calibration}
\end{figure*}

Residual learning augments an existing controller with learned corrective
actions rather than relearning its nominal behavior
\cite{silver2018residual,johannink2019residual}. In humanoid control, Stable
End-Effector Control (SEEC) uses upper-body residuals for end-effector
stabilization \cite{jang2025seec}, while Motion Tracking System with Adaptive
Interface Correction (MOSAIC) applies interface-specific residual adaptation
around a general motion tracker \cite{sun2026mosaic}.

A related line applies residual structures to interaction-specific control.
ResMimic places a task-specific residual over a general motion tracker using
object and contact objectives \cite{zhao2025resmimic}, while Adaptive
Whole-Body Loco-Manipulation (AdaptManip) combines residual manipulation
control with online object-state estimation \cite{zhang2025mst,byrd2026adaptmanip}.
CoorDex is particularly close structurally: it coordinates separate body and
hand priors through a shared residual policy with distinct body and hand
correction heads \cite{li2026coordex}. Its downstream policies, however,
condition on task-specific signals including object-relative geometry,
contact features, goals, and, for staged tasks, task phase. This differs from
the present teleoperation setting, where coordination must be inferred from
human references, initial commands, and robot proprioception without
task-specific object or contact observations.

Context-dependent gating provides a complementary mechanism for restricting
when particular objectives or control behaviors should dominate. Whole-Body
Control with Sequential Contacts (WoCoCo) activates selected rewards
according to predefined contact stages \cite{zhang2024wococo}, Adaptive
Assistive Curriculum Force (A2CF) uses explicit motion phases
\cite{cao2025a2cf}, and the Humanoid Table Tennis Robot (HITTER) restricts
objectives to event-dependent time windows \cite{su2025hitter}. Continuous
geometric gating has also been used to interpolate between locomotion and
manipulation objectives according to robot--target distance
\cite{zhang2025apt,belmontebaeza2025lunar}. These methods span discrete stage- or
event-based activation and continuous geometric modulation, whereas our
gates derive both reward relevance and residual authority from the
human-motion reference.

\section{Multi-Pose Morphology Calibration and Motion Dataset Curation}
\label{sec:calibration-curation}

We construct a motion-reference corpus for Agile One, a humanoid with
29 body degrees of freedom (DoFs) \cite{agilerobots2026agileone}.
Five SEED categories covering locomotion and interaction provide 57,799
SOMA BVH clips; matching their released SMPL counterparts by motion key
retains 52,663 paired candidates. Notably, calibration and curation preserve the
human--robot correspondence throughout.

\subsection{Multi-Pose Morphology Calibration}

We retain NVIDIA SOMA Retargeter's inverse-kinematics (IK) solver \cite{nvidia2026somaretargeter} and develop a multi-pose morphology calibration to compensate for kinematic discrepancies between the source and target morphologies.
We therefore jointly estimate root-frame triaxial scale factors and
effector-local offsets using the seven manually paired human--robot
poses shown in Fig.~\ref{fig:multi-pose-calibration}.

For effector $j$, let $\Delta\mathbf p^h_{j,t}$ be its human root-relative
position, $\mathbf R^h_{r,t}$ the human root orientation, and
$\mathbf S_j=\operatorname{diag}(s_{j,x},s_{j,y},s_{j,z})$ the scale matrix.
The calibrated position target is
\begin{equation}
  \hat{\mathbf p}_{j,t}
  =\mathbf b_{r,t}
  +\mathbf R^h_{r,t}\mathbf S_j(\mathbf R^h_{r,t})^\top
   \Delta\mathbf p^h_{j,t}
  +\hat{\mathbf R}_{j,t}\mathbf d_j,
  \label{eq:triaxial-retargeting}
\end{equation}
where $\mathbf b_{r,t}$ is the separately calibrated root target,
$\hat{\mathbf R}_{j,t}$ incorporates a fixed orientation offset estimated
from the paired poses, and $\mathbf d_j$ is expressed in that effector
frame. Root-frame scaling rotates longitudinal, lateral, and vertical
morphology corrections with the human instead of fixing them to world axes.
With orientation offsets fixed, the seven poses define a linear
least-squares fit of six parameters per effector, uniquely identifiable
when the $21\times6$ design matrix has rank six. The pose set spans
distinct root-relative configurations to improve conditioning; the number
of poses alone does not guarantee identifiability. Bilateral counterparts
are solved jointly with shared scales and side-specific offsets.
The calibrated retargeter processes all 52,663 candidates. Frame-wise
position and orientation checks of pelvis, head, wrist, and ankle targets
remove motions exceeding the configured limits, retaining 46,363 trajectories.
\begin{figure*}[t]
  \centering
  \includegraphics[width=\textwidth]
    {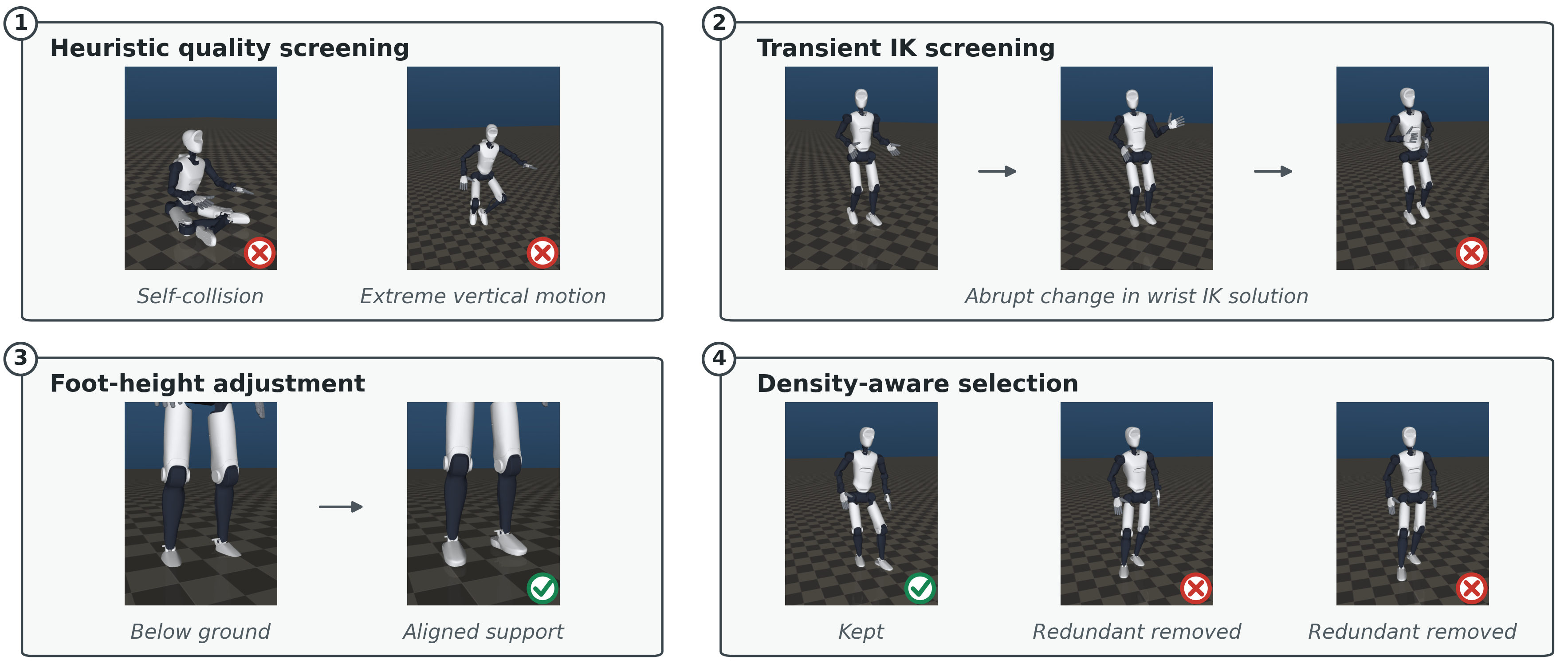}
  \caption{Motion curation through quality screening, transient IK
  screening and repair, foot-height alignment, and density-aware selection.
  Crosses indicate rejection; checks indicate alignment or retention.}
  \label{fig:motion-curation-overview}
\end{figure*}

\subsection{Corpus-Scale Curation}

Geometric target agreement does not ensure a suitable tracking reference:
frame-wise solutions may still contain discontinuities, ground misalignment,
or redundant motion. To address these issues, we propose four complementary curation operations: quality screening, Transient IK repair, foot-height alignment, and density-aware selection, visualized in Fig.~\ref{fig:motion-curation-overview}.

\emph{Quality screening} combines foot sliding, joint velocity,
self-collision, and jerk diagnostics from MuJoCo kinematics and
finite differences, following LIMMT-inspired filtering
\cite{guan2026limmt,todorov2012mujoco}. Aggregating these diagnostics into
a clip-level score retains 31,878 motions.

\emph{Transient IK repair} flags joint changes above $3^\circ$ per frame
at 120~Hz to detect abrupt switches between IK branches. Events affecting
at most two joints in one chain are reconstructed
by cubic interpolation and rechecked; hard events and failed repairs are
rejected. This step repairs 8,537 motions and retains 28,523 in total.

\emph{Foot-height alignment} estimates one constant root-height shift from
reliable double-support frames using foot collision geometry. Low-speed
foot poses near the clip-level minimum height identify support candidates.
The constant shift preserves the trajectory's temporal height variation.
Clips lacking reliable support or retaining penetration above 20~mm are
rejected, leaving 24,821 motions.

\emph{Density-aware selection} prunes dense neighborhoods using a
group-weighted $L_1$ distance over 153 percentile-normalized motion
descriptors covering locomotion, posture, gait, workspace, joint range,
dynamics, periodicity, and symmetry. Removing redundancy, including mirrored
and low-dynamic motions, yields 20,000 references (43.67~h). Across 150
non-constant descriptors, the median retained-to-input percentile-range
ratio is 100.53\%, indicating similar marginal ranges after selection.


\section{Gated Residual Coordination}
\label{sec:gated-residual}
\begin{figure*}[t]
  \centering
  \includegraphics[width=\textwidth]{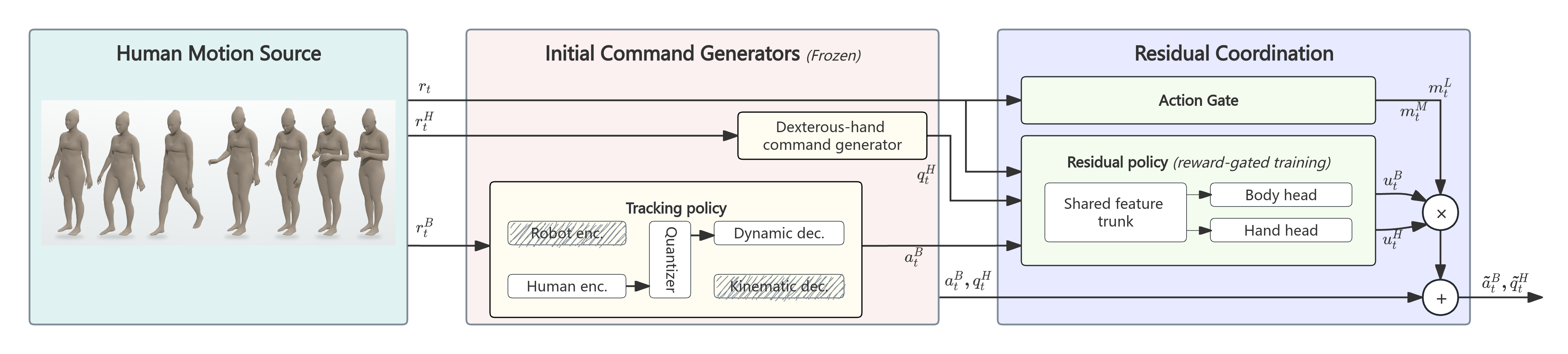}
  \caption{Gated residual coordination. Frozen body and hand generators
produce initial commands. Within SONIC, the hatched robot-motion encoder
and kinematic decoder provide supervision during tracker training and
are inactive during deployment. A shared-trunk, two-head
policy predicts bounded corrections whose authority is modulated by a
human-motion action gate. Geometry-dependent reward gates affect
residual training only.}
  \label{fig:residual-architecture}
\end{figure*}

We train a SONIC-based nominal whole-body tracker on the curated
20,000-motion dataset and freeze it before residual learning.
Fig.~\ref{fig:residual-architecture} summarizes the resulting
body--hand coordination framework.
The residual coordinates frozen body and hand command generators without
object-state or contact observations.
An action gate modulates correction authority during training and inference;
separate geometry-dependent reward gates weight interaction objectives
only during training. This separation controls both how strongly the
policy may alter nominal commands and when particular geometric relations
should shape learning.

\subsection{Base Tracker Training}
\label{sec:base-tracker-training}

The SONIC-based nominal tracker is trained for 30,000 iterations of proximal policy optimization
(PPO) \cite{schulman2017ppo} on eight RTX PRO 6000 GPUs
(approximately 1.1k GPU-hours). The curriculum increases default-pose
initialization, dynamics randomization, root-velocity perturbations, and
motion/joint-limit regularization. The trained tracker is frozen before
learning command-level coordination.

\subsection{Residual Command Interface}

Let $\mathbf h_t$ combine SMPL body motion \cite{loper2015smpl} and
SMPL-X-compatible hand keypoints \cite{pavlakos2019smplx}.
At 50~Hz, the frozen tracker outputs a normalized body action
$\mathbf a^B_t=\pi_B(\mathbf o^B_t,\mathbf h_{t:t+9})\in\mathbb R^{29}$.
A DexPilot-style retargeter \cite{handa2019dexpilot} maps wrist-local
human hand keypoints to joint-angle targets $\mathbf q^H_t\in\mathbb R^{32}$,
with 16 active joints per hand.
The residual actor receives both commands, current root-local body and
wrist-local hand references, and a proprioceptive history of projected
gravity, base angular velocity, joint states, and previous merged commands.
These inputs expose both the intended motion and the robot's response to
the independently generated commands.
A shared trunk feeds 29-dimensional body and 32-dimensional hand heads,
producing $\mathbf u^B_t$ and $\mathbf u^H_t$. Their zero-initialized final
projections recover direct composition in the initial deterministic mean.
Body corrections modify normalized tracker actions, while hand corrections
modify physical joint-angle targets:
\begin{equation}
\begin{aligned}
  \tilde{\mathbf a}^B_t
  &=\mathbf a^B_t+\mathbf m^B_t\odot\mathbf s^R_B
    \odot\tanh(\mathbf u^B_t),\\
  \tilde{\mathbf q}^H_t
  &=\mathbf q^H_t+m^M_t\mathbf s^R_H\odot\tanh(\mathbf u^H_t).
\end{aligned}
\label{eq:residual-commands}
\end{equation}
Here, $\mathbf s^R_B,\mathbf s^R_H$ specify fixed residual bounds;
$\mathbf m^B_t$ assigns $m^L_t$ to legs and $m^M_t$ to waist and arms,
including wrists. Leg corrections are more tightly bounded, and head
corrections are disabled. The shared representation supports coordination
across streams, while separate heads and scales retain their distinct
output spaces. Setting all bounds to zero exactly recovers
direct composition through the same command path.

\subsection{Human-Motion Action Gate}

The gate uses 25 reference frames: 15 past, the current frame, and nine
future samples already supplied by SONIC's 180~ms buffer, adding no
further lookahead. Manipulation features capture root-to-wrist workspace,
wrist activity, finger activity, and bimanual proximity; locomotion features
capture leg range, vertical foot swing, and bilateral alternation.
Proximity increases as the wrists approach; alternation derives from
left--right motion correlation. Range features use $P_{90}-P_{10}$ statistics.
Features are normalized to $[0,1]$ with fixed bounds and aggregated with
fixed weights. A smoothstep over $[0,0.60]$ and exponential smoothing
with coefficient 0.25 produce
the scores $\bar g^M_t,\bar g^L_t\in[0,1]$, which yield
$m^M_t=0.5+0.5\bar g^M_t$ and $m^L_t=0.5+0.5\bar g^L_t$.
The nonzero floor retains support correction during manipulation and
upper-body/hand correction during locomotion; the gate modulates authority
continuously rather than selecting a single active controller.

\subsection{Geometry-Gated Learning}

We compare task-space relations because human and robot joint
configurations differ in limb lengths, hand dimensions, and reachable
workspaces. Tracking terms compare root orientation and horizontal position,
wrist--root orientation and direction,
within-hand thumb-to-fingertip distances, corresponding inter-hand
fingertip distances, and the inter-wrist relative transform. Fixed frame
rotations align robot and human wrist conventions. Position and distance
terms use mean-squared errors with rewards
$\rho_k(E)=\exp(-E/\sigma_k^2)$, where $\sigma_k$ controls tolerance;
rotations use
geodesic error on $\mathrm{SO}(3)$, directions use cosine error, and
inter-wrist position and rotation are normalized separately.
The objective is
\begin{equation}
  r_t=\Delta t\left[
    \sum_k w_kG_{k,t}\rho_k(E_{k,t})
    +\sum_j\lambda_jc_{j,t}\right],
  \label{eq:residual-reward}
\end{equation}
where $\Delta t=0.02$~s, $E_{k,t}$ denotes relation error, $w_k$ the
tracking weight, and $c_{j,t}\geq0$ a cost with $\lambda_j\leq0$.
Logistic close gates downweight inter-hand fingertip and inter-wrist
objectives as human-reference distances increase through 0.10--0.20~m
and 0.20--0.40~m, respectively; a far gate increases wrist-direction
weight over 0.20--0.40~m. These gates emphasize relative hand geometry
during close interaction and wrist direction when the arm extends.
Gate distances come exclusively from the human reference, so their
evaluation requires neither object states nor contact labels.
Other terms use $G_{k,t}=1$.
Costs penalize joint motion, soft-limit use, self-collision, residual
magnitude, lower-body correction, and root/wrist temporal changes.
We train on 2,304 GRAB motions, including mirrors, and 1,006 PICO
references \cite{taheri2020grab,pico2026enterprise,zhao2026xrobotoolkit}
using PPO for 5,000
iterations on eight RTX PRO 6000 GPUs. GRAB supplies synchronized
body--hand interaction motion, while PICO replay matches the intended
teleoperation input format. Training progressively increases default-pose
initialization, domain randomization, and motion/collision/residual
regularization while keeping the reference mixture and objectives fixed.
An asymmetric critic additionally accesses future human references and
local hand geometry during training; both command generators remain frozen.

\section{Experiments}

We evaluate four aspects of the system: retargeting, SONIC tracker adaptation,
residual ablation, and preservation of tracking capability.

\subsection{Retargeting Calibration}

We compare the scalar single-pose baseline with the proposed triaxial
seven-pose calibration by applying both configurations to the same
52,663 paired motion candidates. Calibration parameters are estimated
from the respective manually paired pose sets. Solver iterations, validation rules, temporal
settings, and post-processing are fixed; the two configurations differ only
in their calibration procedure. Position errors are measured after inverse
kinematics relative to the calibrated targets generated by the corresponding
configuration and are aggregated frame-wise over complete trajectories.
Table~\ref{tab:icra-retargeting-comparison} reports the resulting errors.

\begin{table}[h]
  \centering
  \caption{Comparison of post-IK positional errors.}
  \label{tab:icra-retargeting-comparison}
  \scriptsize
  \begin{tabular*}{\columnwidth}{@{\extracolsep{\fill}}lrrr@{}}
    \toprule
    Method & Root (mm) & Wrist (mm) & Ankle (mm) \\
    \midrule
    Scalar, 1 pose
    & 23.16
    & 76.59
    & \textbf{1.67} \\
    Triaxial, 7 poses
    & \textbf{20.48}
    & \textbf{54.82}
    & 2.03 \\
    \bottomrule
  \end{tabular*}
\end{table}

The triaxial seven-pose calibration reduces post-IK root and wrist target
errors by 11.6\% and 28.4\%, respectively. Ankle target error increases by
0.36~mm, from 1.67 to 2.03~mm.

\subsection{SONIC Tracker Evaluation}

We evaluate SONIC G1 Released, our SONIC G1 Comparable policy, and SONIC AO on
8,802 held-out AMASS motions with GRAB excluded
\cite{mahmood2019amass}. Each motion has ten observation-corrupted replicas.
Success runs apply root and distal-height terminations, with positional
thresholds scaled by robot mesh height and orientation thresholds unchanged.
Tracking runs retain all frames to motion end. Reported position errors are
normalized by robot height, while orientation errors are measured in radians.

SONIC G1 Comparable and SONIC AO use the same training source corpus and
budget, corresponding to approximately 40~h of motion data and 1.1k GPU-hours.
SONIC G1 Released uses its broader internal training corpus
(approximately 700~h and 21k GPU-hours \cite{luo2025sonic}) and is included as
a non-data-matched reference. Table~\ref{tab:icra-sonic-tracking} reports
motion-equal errors from the tracking runs.

\begin{table}[!ht]
  \centering
  \caption{Base SONIC tracking on held-out AMASS.}
  \label{tab:icra-sonic-tracking}
  \scriptsize
  \begin{tabular*}{\columnwidth}{@{\extracolsep{\fill}}lrrr@{}}
    \toprule
    Policy
    & Success (\%)
    & Root XY / ori.
    & Body pos. / ori. \\
    \midrule
    SONIC G1 Released
    & 98.41
    & 0.2196 / 0.1447
    & 0.0288 / 0.2381 \\
    \midrule
    SONIC G1 Comparable
    & \textbf{89.95}
    & \textbf{0.2291} / \textbf{0.1662}
    & \textbf{0.0392} / 0.3449 \\
    SONIC AO
    & 89.42
    & 0.2346 / 0.2000
    & 0.0407 / \textbf{0.2915} \\
    \bottomrule
  \end{tabular*}
\end{table}

SONIC AO achieves nearly the same height-scaled success as the data- and
budget-matched G1 policy. Their tracking errors are also comparable:
SONIC G1 Comparable has lower root XY, root-orientation, and body-position
errors, while SONIC AO has lower body-orientation error. SONIC G1 Released
achieves the highest success rate and the lowest error in each reported
tracking measure.
\begin{figure*}[t]
  \centering
  \includegraphics[width=\textwidth]
  {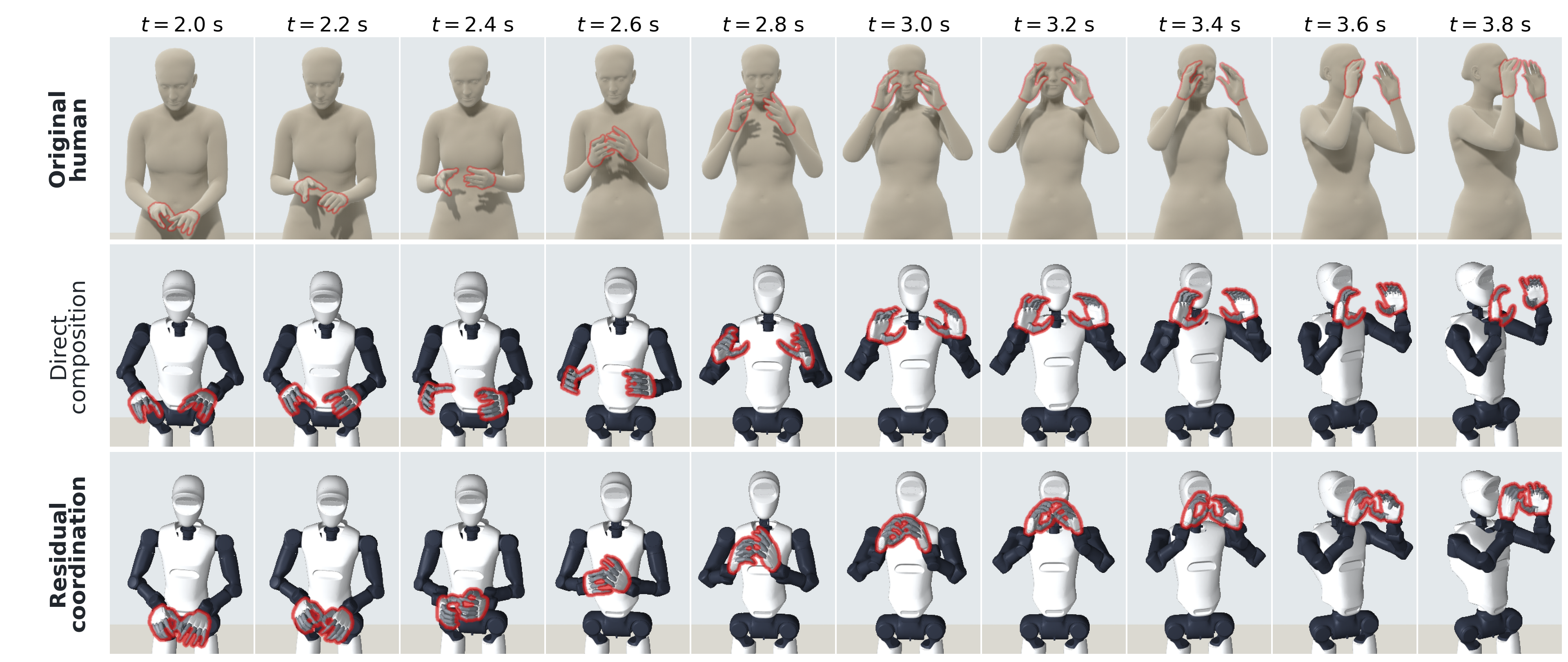}
  \caption{Synchronized comparison on an unseen GRAB motion. Rows show the
  human reference, direct composition, and residual coordination.}
  \label{fig:heldout-binocular-qualitative}
\end{figure*}

\begin{table*}[t]
  \centering
  \caption{Held-out GRAB ablation geometry and raw residual saturation.}
  \label{tab:grab-geometry}
  \begin{tabular*}{\textwidth}{@{\extracolsep{\fill}}lrrrrrrr@{}}
    \toprule
    Condition & Root XY(m) & Root ori.(deg) & Wrist dist.(cm) & Wrist rot.(deg) &
    Inter-hand tips(cm) & Thumb--tips(cm) & Raw sat.(\%) \\
    \midrule
    Residual Off & 0.145 & 5.01 & 10.26 & 29.15 & 11.61 & 4.33 & N/A \\
    \textbf{Full Residual} & 0.110 & 4.38 & 4.62 & 17.73 & 5.07 & 2.28 & 1.40 \\
    Constant Reward Gate & 0.107 & 4.08 & 3.35 & 16.25 & 4.02 & 2.21 & 1.63 \\
    Constant Action Gate & 0.129 & 4.41 & 4.94 & 16.76 & 4.94 & 2.35 & 1.92 \\
    Single Head & 0.123 & 4.48 & 4.53 & 17.32 & 4.97 & 2.29 & 2.10 \\
    Upper-Body Only & 0.153 & 4.86 & 4.57 & 17.55 & 5.04 & 2.39 & 1.70 \\
    \bottomrule
  \end{tabular*}
\end{table*}

\subsection{Residual Ablation}

The residual benchmark uses 50 held-out GRAB motions that are absent from the
2,304 GRAB motions used during residual training. It covers all ten subjects,
26 object classes, and 15 motion intents, with ten motions in each of five
predefined regimes: Low Activity, High Locomotion, High Manipulation,
Close Bimanual, and Gate Transition. Each motion uses ten matched
replicas. All conditions share the articulated-hand plant, frozen command
generators, human references, observation corruption, and replica seeds.
Replica metrics are averaged within each motion before motion-equal
aggregation.

We compare Residual Off, Full Residual, Constant Action Gate, Constant Reward Gate, a
capacity-matched Single Head, and Upper-Body Only. Residual Off follows the
Full Residual inference path with all residual scales set to zero and requires no
separate training. All learned variants use the same GRAB--PICO reference
mixture, training budget, and optimization schedule. All conditions reach
motion end with 100\% root-tracking success, so continuous geometric errors
provide the primary comparison.

Constant Action Gate fixes the upper-body/hand and leg multipliers to 0.7703
and 0.5967. These are Full Residual's frame-weighted training-corpus means, so mean
nominal authority is matched while temporal variation is removed; learned
applied corrections may still differ. Constant Reward Gate fixes the
inter-hand, inter-wrist, and wrist-direction gates to their corresponding
means, 0.0091, 0.1122, and 0.9443. Single Head uses one 61-D decoder with
1,091,533 parameters, within 0.37\% of the routed dual decoder. Upper-Body Only
sets the applied scales of the 12 leg residuals and their L2 penalty to zero,
while retaining waist, arm, wrist, and 32 active-hand residuals. The two head
residuals are disabled in all conditions.

Raw residual saturation is the motion-equal mean fraction of active residual
channel--time entries satisfying $|\tanh(u_i)|>0.99$ before gating and scaling.
Channels with a nonzero applied scale enter the denominator: 59 for all learned
conditions except Upper-Body Only, which uses 47. Residual Off is reported as
N/A because it applies no residual channels.

Figure~\ref{fig:heldout-binocular-qualitative} provides a synchronized
qualitative comparison on a held-out bimanual motion.

Table~\ref{tab:grab-geometry} reports root and body--hand coordination errors.
Relative to direct composition, Full Residual reduces root XY error by 24.2\% and all four reported wrist and fingertip geometry errors by 39.2--56.3\%.

The ablations expose different geometric trade-offs. Constant Reward Gate
achieves the lowest aggregate coordination errors. In Close-Bimanual windows,
Full Residual reduces wrist-distance error from 6.54 to 6.11~cm and
wrist-rotation error from 23.57$^\circ$ to 22.33$^\circ$ relative to Constant
Reward Gate. Outside these windows, its raw saturation is lower, at 0.96\%
versus 1.62\%. In Close-Bimanual windows, the corresponding rates are 3.15\%
and 1.67\%, respectively.
Constant Action Gate increases root XY error from 0.110 to 0.129~m under the
matched mean multipliers, consistent with a benefit from state-dependent
authority allocation. Removing leg residual authority further increases root
XY error to 0.153~m. Single Head increases raw residual saturation from
1.40\% to 2.10\%, without a consistent geometric advantage over
Full Residual.

\subsection{Whole-Body Tracking Preservation}

Using the same AMASS protocol, we next separate the effect of the articulated
hand plant from that of residual correction. The base articulated-hand
condition applies the frozen SONIC AO policy while continuously commanding a
neutral hand posture. The residual condition uses the same plant and neutral
hand references, differing only by the residual policy. Errors are measured
against the retargeted robot reference. Table~\ref{tab:amass-preservation} reports the comparison. As in the preceding
AMASS evaluation, position errors are normalized by robot height and
orientation errors are reported in radians.

\begin{table}[t]
  \centering
  \caption{AMASS tracking-preservation diagnostic.}
  \label{tab:amass-preservation}
  \scriptsize
  \begin{tabular*}{\columnwidth}{@{\extracolsep{\fill}}lrrr@{}}
    \toprule
    Condition
    & Success (\%)
    & Root XY / ori.
    & Body pos. / ori. \\
    \midrule
    Base, fixed hand
    & 89.42
    & 0.2346 / 0.2000
    & 0.0407 / 0.2915 \\
    \midrule
    Base, articulated hand
    & 89.03
    & \textbf{0.2471} / 0.2089
    & \textbf{0.0426} / \textbf{0.2971} \\
    Base + residual
    & \textbf{89.29}
    & 0.2509 / \textbf{0.2020}
    & 0.0451 / 0.3190 \\
    \bottomrule
  \end{tabular*}
\end{table}

On the same articulated-hand plant, enabling the residual changes
height-scaled success from 89.03\% to 89.29\%. Root-orientation error decreases
from 0.2089 to 0.2020~rad, while root XY, body-position, and body-orientation
errors increase modestly.

\section{Discussion and Limitations}

The retargeting comparison shows lower post-IK root and wrist errors
relative to each configuration's calibrated targets, with a small
increase in ankle error. The resulting SONIC AO tracker achieves
comparable performance to the data- and budget-matched G1 policy, supporting
the feasibility of adapting a generalist tracker to a new humanoid embodiment.
The released G1 policy remains a stronger non-data-matched reference under a
substantially larger training scale.

The held-out GRAB results support the central claim that bounded residual
correction improves body--hand geometric coordination over direct composition.
The two gates serve complementary roles: the reward gate localizes when
interaction-specific objectives shape learning, whereas the action gate
modulates correction authority according to motion context. The full model
therefore represents a trade-off among bimanual coordination, root execution,
and residual use rather than minimizing every aggregate metric.

The same-plant AMASS comparison indicates that the residual preserves broad
whole-body tracking rather than uniformly improving robot-reference tracking.
Several limitations remain. The policy lacks explicit object-state,
contact-force, and tactile feedback. Evaluation is limited to simulation, one
humanoid embodiment, and one training seed, and the action gate relies on the
base system's existing 180~ms reference buffer.

\section{Conclusion}

We presented an object-agnostic gated residual framework for coordinating modular whole-body and dexterous-hand command streams. To establish the nominal body stream on Agile One, we introduced a multi-pose morphology-calibration method that jointly estimates triaxial scales and effector-local offsets, together with staged motion dataset curation for training a SONIC-based whole-body tracker. The residual policy keeps this tracker and the hand retargeter frozen while applying bounded, motion-conditioned corrections to their commands. Simulation results demonstrate improved bimanual wrist and fingertip geometry while maintaining comparable whole-body tracking ability.

Future work will investigate tactile feedback for contact-aware residual
coordination and neural hand retargeting for improved human--robot hand
correspondence. We will also evaluate more extensively pretrained
whole-body trackers to assess how stronger nominal tracking affects
residual coordination and generalization across a broader range of motions.

\bibliographystyle{IEEEtran}
\bibliography{residual_refs}

\end{document}